\documentclass[letterpaper, 10 pt, conference]{ieeeconf}  % Comment this line out if you need a4paper

\IEEEoverridecommandlockouts                              % This command is only needed if 
\usepackage{graphicx} % for pdf, bitmapped graphics files
\usepackage{mathtools}

\usepackage{amsmath} % assumes amsmath package installed
\usepackage{textcomp}
\usepackage{tabularx,booktabs}
\usepackage{color}
\usepackage[pdftex,dvipsnames]{xcolor}
\usepackage{array}
\usepackage{algorithm} 
\usepackage{algpseudocode} 
\usepackage{subcaption} %subfigure

\newcolumntype{P}[1]{>{\centering\arraybackslash}p{#1}}
\newcolumntype{C}{>{\centering\arraybackslash}X} % centered version of "X" 

\usepackage{tikz}

\title{\LARGE \bf
Patch-DrosoNet: Classifying Image Partitions With Fly-Inspired Models For Lightweight Visual Place Recognition
}

\author{Bruno Arcanjo$^{1}$, Bruno Ferrarini$^{1}$, Michael Milford$^{2}$, Klaus D. McDonald-Maier$^{1}$ and Shoaib Ehsan$^{1, 3}$% <-this % stops a space
\thanks{}% <-this % stops a space
\thanks{$^{1}$B. Arcanjo, B. Ferrarini, K. D. McDonald-Maier and S. Ehsan are with the School of Computer Science and Electronic Engineering, University of Essex, United Kingdom {\tt\small (email: bq17319@essex.ac.uk; bferra@essex.ac.uk; kdm@essex.ac.uk; sehsan@essex.ac.uk)}}%
\thanks{$^{2}$M. Milford is with the School of Electrical Engineering and Computer Science, Queensland University of Technology, Brisbane, QLD 4000, Australia
        {\tt\small (email: michael.milford@qut.edu.au)}}%
\thanks{$^{3}$S. Ehsan is also with the school of Electronics and Computer Science, University of Southampton, United Kingdom \tt\small{(email: s.ehsan@soton.ac.uk)}}%
}
\begin{document}

\maketitle
\thispagestyle{empty}
\pagestyle{empty}

%%%%%%%%%%%%%%%%%%%%%%%%%%%%%%%%%%%%%%%%%%%%%%%%%%%%%%%%%%%%%%%%%%%%%%%%%%%%%%%%
\begin{abstract}
Visual place recognition (VPR) enables autonomous systems to localize themselves within an environment using image information. While Convolutional Neural Networks (CNNs) currently dominate state-of-the-art VPR performance, their high computational requirements make them unsuitable for platforms with budget or size constraints. This has spurred the development of lightweight algorithms, such as DrosoNet, which employs a voting system based on multiple bio-inspired units.
In this paper, we present a novel training approach for DrosoNet, wherein separate models are trained on distinct regions of a reference image, allowing them to specialize in the visual features of that specific section. Additionally, we introduce a convolutional-like prediction method, in which each DrosoNet unit generates a set of place predictions for each portion of the query image. These predictions are then combined using the previously introduced voting system.
Our approach significantly improves upon the VPR performance of previous work while maintaining an extremely compact and lightweight algorithm, making it suitable for resource-constrained platforms.

\end{abstract}

\section{Introduction \& Background}
\label{intro}

Visual place recognition (VPR) is an essential component of mobile robotics, as it allows the system to localize itself in the runtime environment using only image data \cite{ref:vpr-survey}. The affordability and variety of camera sensors makes VPR localization particularly attractive for hardware restricted robotic platforms, which are common in mobile robotics \cite{ref:res_hardware_1, ref:res_hardware_2}. Nevertheless, VPR is a complicated task and proposed solutions must deal with several visual challenges. The same place can appear vastly different when visited under different illumination \cite{ref:illu_changes}, seasonal weather conditions \cite{ref:season_changes}, viewpoints \cite{ref:pov_changes} and elements entering or leaving the scene \cite{ref:dyna_changes}. As previously mentioned, mobile robotic platforms often operate with low-end hardware, making computational cost yet another important consideration when designing VPR techniques.

The importance of VPR and its variety of challenges has resulted in a growing number of approaches being proposed in the literature \cite{masone_survey_2021}. Some of the first successful solutions \cite{ref:fabmap} relied on handcrafted local feature descriptors, such as Scale-Invariant Feature Transform (SIFT) \cite{ref:sift} and Speeded-up Robust Features (SURF) \cite{ref:surf}, to build a viewpoint-robust map of the environment. Despite their strong resilience to viewpoint changes, local feature based approaches are susceptible to strong appearance changes. Whole-image descriptors, like Histogram Oriented Gradients (HOG) \cite{ref:hog} have also been employed in VPR \cite{ref:mcmanus2014scene}. Recently, state-of-the-art VPR performance has been achieved by using Convolutional Neural Network (CNN) based approaches \cite{ref:cnn_for_vpr1}, as features from the inner layers of trained CNNs have been shown to significantly improve VPR performance \cite{ref:labaucs}. Several CNN-based techniques \cite{ref:hybridasmosnet, ref:netvlad, hou2015convolutional} have thus been successfully employed for performing VPR.

The impressive VPR performance offered by CNN-based approaches comes with the significant downside of a high-computational cost, often demanding powerful graphical processing units (GPUs) to be ran in real time \cite{ref:res_hardware_2}. This shortcoming makes these top-performing techniques unusable for hardware-restricted platforms and several lightweight VPR algorithms have hence emerged to address it. HOG has been shown to be a fast VPR descriptor if used with suitable image sizes. CoHOG \cite{ref:cohog} makes use of the efficient HOG to encode high-entropy regions of an image, improving resilience to viewpoint changes but significantly increasing place matching computation times. \cite{ref:patch-netvlad} is also a region-based method, adapting the high-performing and costly NetVLAD \cite{ref:netvlad} descriptor. CALC \cite{ref:calc} presents itself as a train-free, lightweight CNN model, capable of competitive real-time VPR performance. Bio-inspired VPR approaches attempt to replicate the efficient localization abilities of small animals, resulting in algorithms such as \cite{ref:ratslam, ref:flynetcann}. Recently, a lightweight VPR voting system \cite{voting_arcanjo} based on multiple units, each dubbed DrosoNet, inspired by the odour processing abilities of the fruit fly \cite{ref:droso_vpr} has been proposed. The system capitalizes on the inherit randomness of the initialization and training process of individual units, where different models might specialize better or worse on different visual features. The combination of multiple units via the voting mechanism attempts to eliminate the weak spots of some units with the strengths of others. However, since each DrosoNet is trained on the entirety of each reference image, the sources of feature specialization are minimal.

In this work, we propose a novel, region-based approach to train individual DrosoNets coupled with a convolution-like prediction mechanism. By training different DrosoNets on different sections of the reference images, we introduce another source of variation between different units, rather than just relying on random initialization and training. Furthermore, the matching mechanism makes each unit place a prediction for each region in the query image, providing more information during voting. Our approach significantly improves VPR performance when compared to previous work, while retaining the lightweight capabilities of the technique.

\section{Method}
\label{method}

In this section, we provide implementation details of our proposed Patch-DrosoNet algorithm. Firstly, a quick overview of the basic functionality of DrosoNet is given. Then, we explain the processes of splitting the images, training the DrosoNets on different image patches and place-matching at runtime.

\subsection{DrosoNet Usage}
As explained in \cite{voting_arcanjo}, DrosoNet is a compact and fast neural network image classifier inspired by Drosophila Melanogaster, where each  of the total $N$ places is a different class. DrosoNet works as a classification function:
\begin{equation}
    D(i) = S_i
\end{equation}

\noindent where $i$ is the $32\times64$ input image and $S_i$ is the output score vector of $N$ elements, where each score corresponds to one of the reference places. The class obtaining the highest score in $S$ is output as the place prediction.

While DrosoNet is a fast algorithm, its standalone VPR performance is lower than more computationally intensive techniques. Moreover, due to the randomness of its initialization and training, different DrosoNets exhibit high variance in their performance with different visual conditions. Combining multiple DrosoNets was hence proposed as a measure to improve overall VPR performance, relying only the native stochastic behaviour of the models for differentiation \cite{voting_arcanjo}.

Our novel approach is to train multiple groups of DrosoNets on different patches of the same image, specializing each group on the features of each image region.

\subsection{Image Splitting}
The image splitting takes an input image $i$ and returns a grid of patches of dimensions $r\times c$, where $r$ is the number of rows and $c$ the number of columns in the grid. Since DrosoNet operates with $32\times64$ images, we first resize the image to $32r\times 64c$. Within the same dataset, all images are split in the same fashion, and a group of DrosoNets is assigned to each patch.

\subsection{Training}
One DrosoNet group is assigned to each patch of the split reference images and it is trained only on those particular regions. The goal is for each DrosoNet group to specialize on the visual features of a region of the reference images. The total number of DrosoNets $T$ in the algorithm is hence riven by 

\begin{equation}
    T = rcz
\end{equation}

\noindent where $z$ is an hyperparameter denoting the number of DrosoNets to use per patch.

\subsection{Matching}

\begin{figure}[thpb]
\vspace*{1ex}
\centering
\includegraphics[width=0.75\columnwidth]{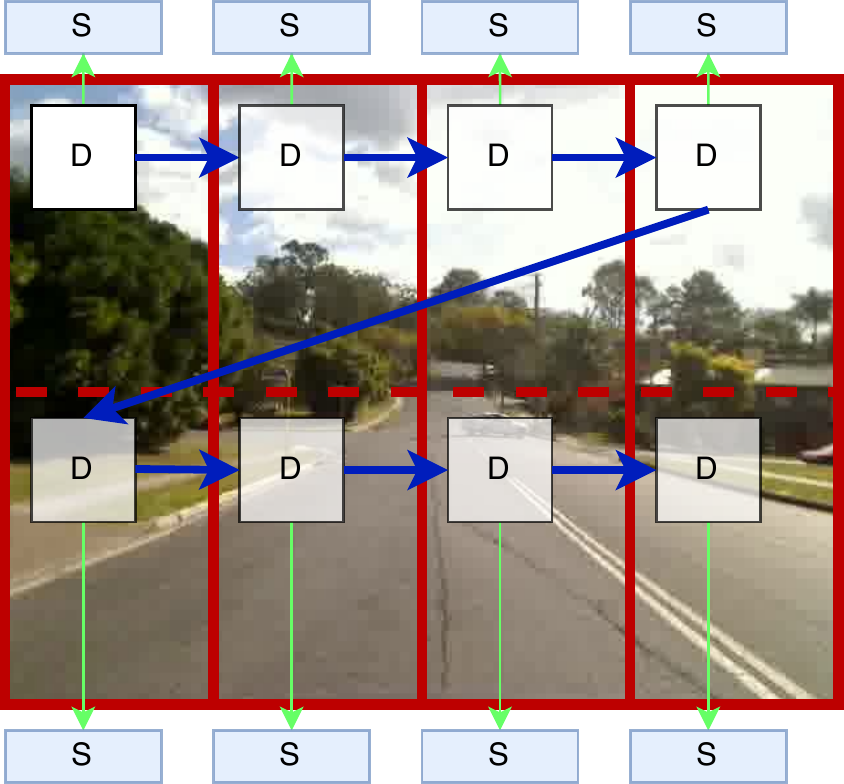}
\caption{Matching process on $2\times4$ grid.}
\label{fig:matching}
\end{figure}

At matching time, each DrosoNet evaluates all sections of the query image, regardless of what patch it was trained on. This process generates $rc$ score vectors per query image per DrosoNet D. Fig. \ref{fig:matching} provides an example of the matching process for a split grid of $2\times4$ regions, yielding a total of $8$ score vectors per DrosoNet.

The total number of DrosoNet function calls per query image, $C$, and consequently also the total number of score vectors, can be computed by 

\begin{equation}
    C = Trc
\end{equation}

\noindent Note how the number of regions factor $rc$ is squared, and it is therefore important to use a low number of patches to preserve computational efficiency. Finally, all the $C$ score vectors produced by the $T$ different DrosoNets are merged using the same voting system as in \cite{voting_arcanjo}, producing a final score vector from which the final prediction can be extracted.

The exhaustive approach to matching is required, as the visual features of a reference region might be partially present in a different patch of the query due to viewpoint changes. 

\section{Experiments}
In this section, we provide details on our experimental setup, such as model configurations and datasets. We then present and discuss our results in terms of VPR performance and computational efficiency.

\subsection{Setup}
We use the datasets in Table \ref{tab:datasets}, allowing for a margin of error of 1 frame for Nordland Winter and Fall \cite{ref:nord} and of 2 frames for Day-Right and St. Lucia \cite{stlucia}.

We compare our proposed approach against other lightweight techniques such as CoHOG, CALC and the established Voting system. For the first two techniques, we use the implementations provided in \cite{ref:vpr_bench}, while for Voting we use the settings in \cite{voting_arcanjo}. For Patch-DrosoNet, we use the same DrosoNet configuration as in \cite{voting_arcanjo}, with image-grids and DrosoNets per patch as detailed in Table \ref{tab:datasets}.

\newcolumntype{M}[1]{>{\centering\arraybackslash}m{#1}}
\begin{table}[t]
  \centering
  \vspace{5px}
  \caption{Dataset Details}
  \label{tab:datasets}%
    \begin{tabular}{M{1cm}M{1.5cm}M{1.5cm}M{1.5cm}M{1cm}}
    \toprule
Dataset  & Condition   & Image-Grid  & DrosoNets Per Patch & Number of Images \\
\midrule
\midrule
Nordland Winter  & Extreme seasonal    & 3x1   & 16      &   1000      \\
\midrule
Nordland Fall        & Moderate seasonal   & 3x1  &  4       &  1000   \\
\midrule
Day-Right & Outdoor Lateral Shift & 1x3     & 8     &  200 \\
\midrule
St. Lucia   & Daylight; Dynamic Elements &   4x2   &    8   &   1100  \\
\bottomrule
\end{tabular}
\end{table}%

\subsection{Results}
\label{results}

\begin{figure*}[!t]
	\centering
	\begin{subfigure}[b]{0.48\textwidth}
		\centering
		\includegraphics[width=0.8\linewidth]{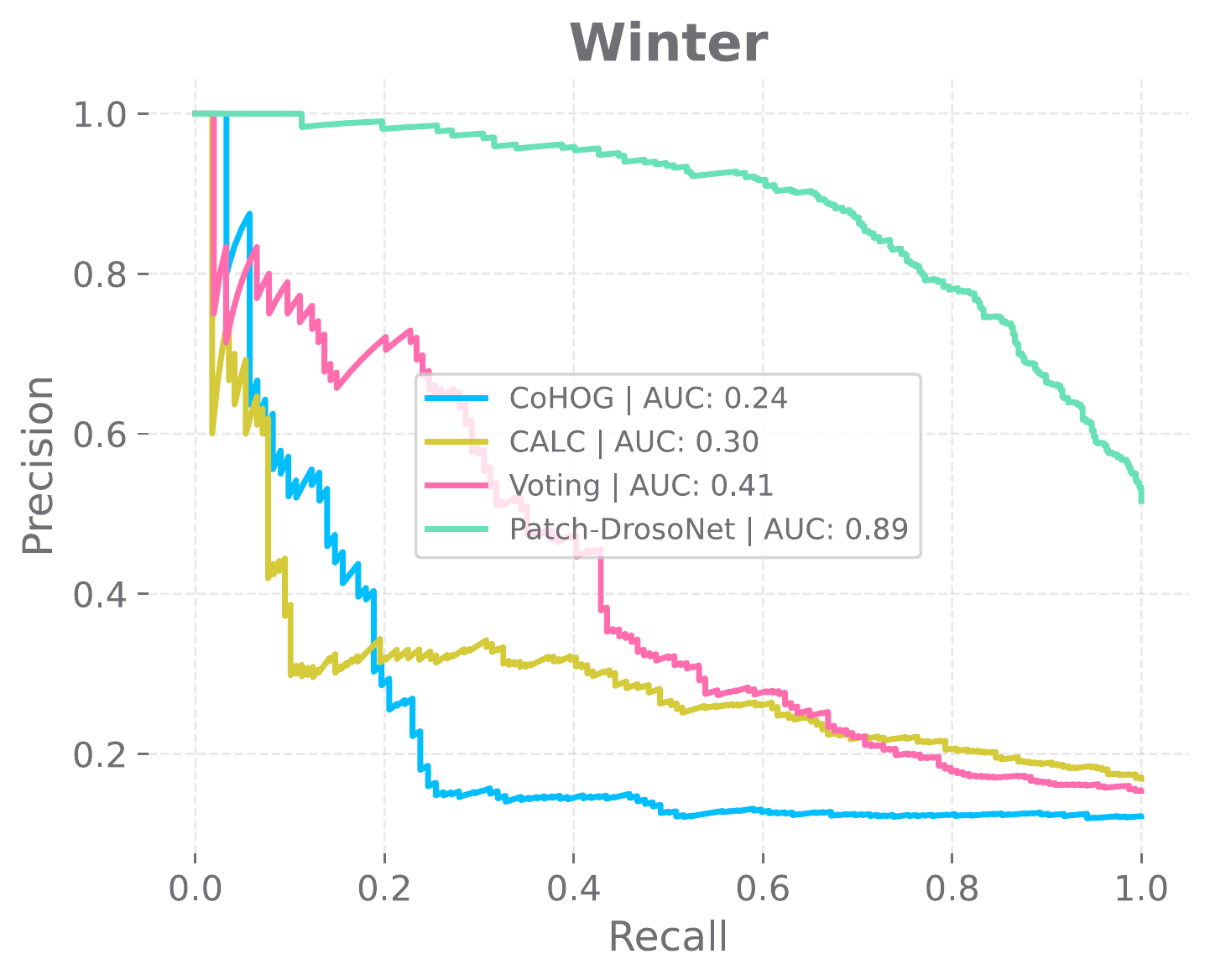}
		\caption{}
		\label{fig:pr_curves:A}
	\end{subfigure}
	\hfill
	\begin{subfigure}[b]{0.48\textwidth}
		\centering
		\includegraphics[width=0.8\linewidth]{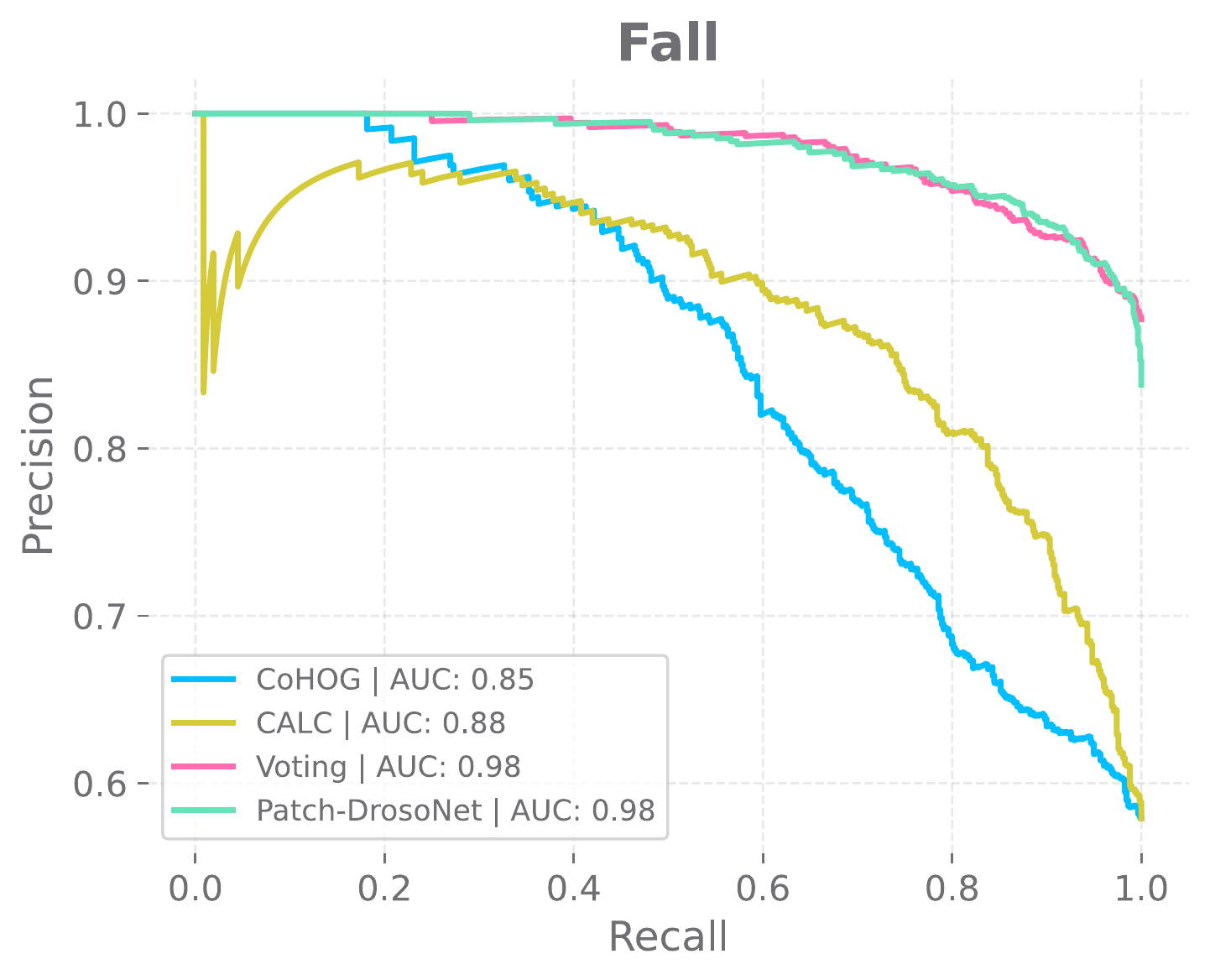}
		\caption{}
		\label{fig:pr_curves:B}
	\end{subfigure}
	\hfill	
	\begin{subfigure}[b]{0.48\textwidth}
		\centering
		\vspace{2ex}
		\includegraphics[width=0.8\linewidth]{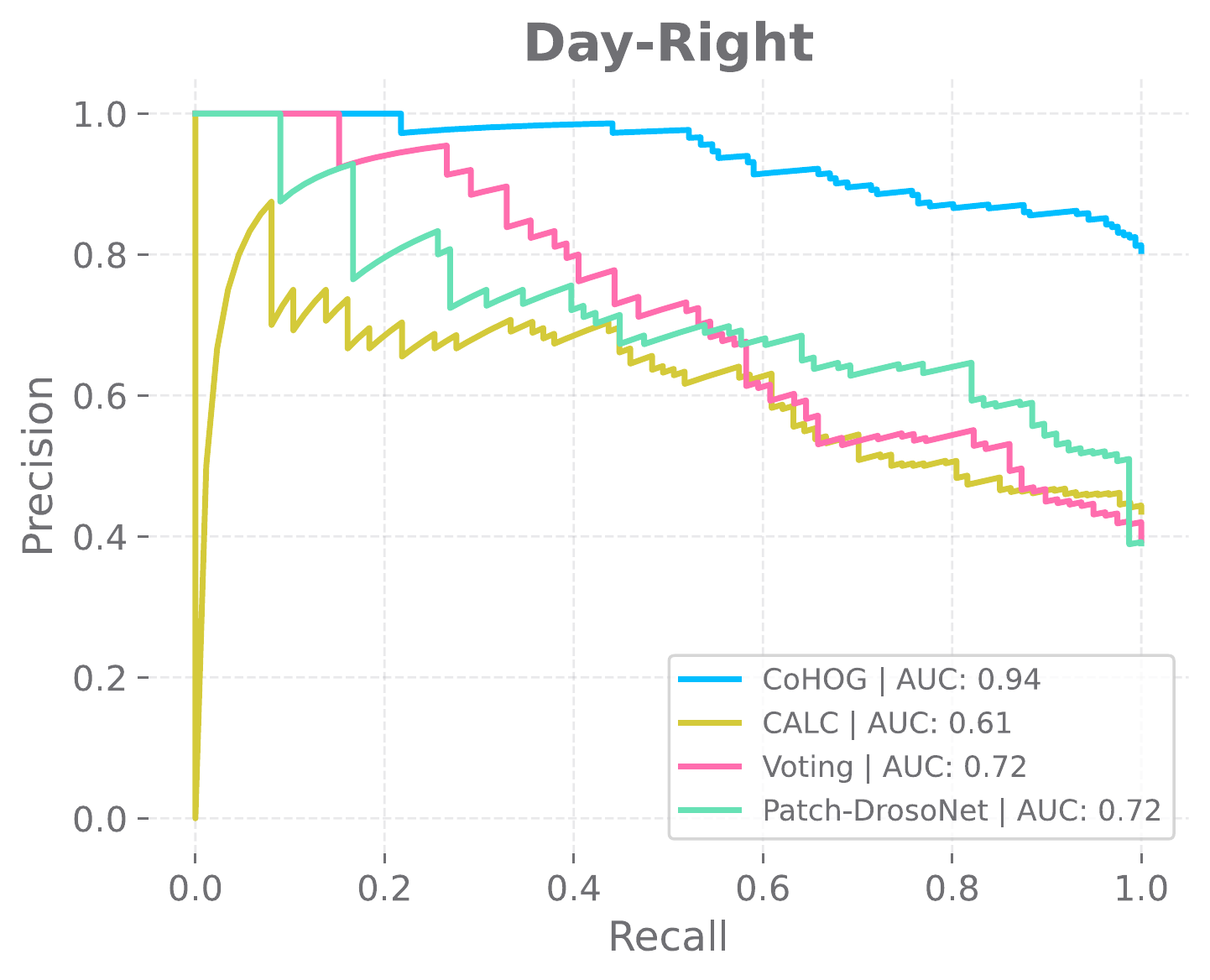}
		\caption{}
		\label{fig:pr_curves:C}
	\end{subfigure}	
        \begin{subfigure}[b]{0.48\textwidth}
		\centering
		\includegraphics[width=0.8\linewidth]{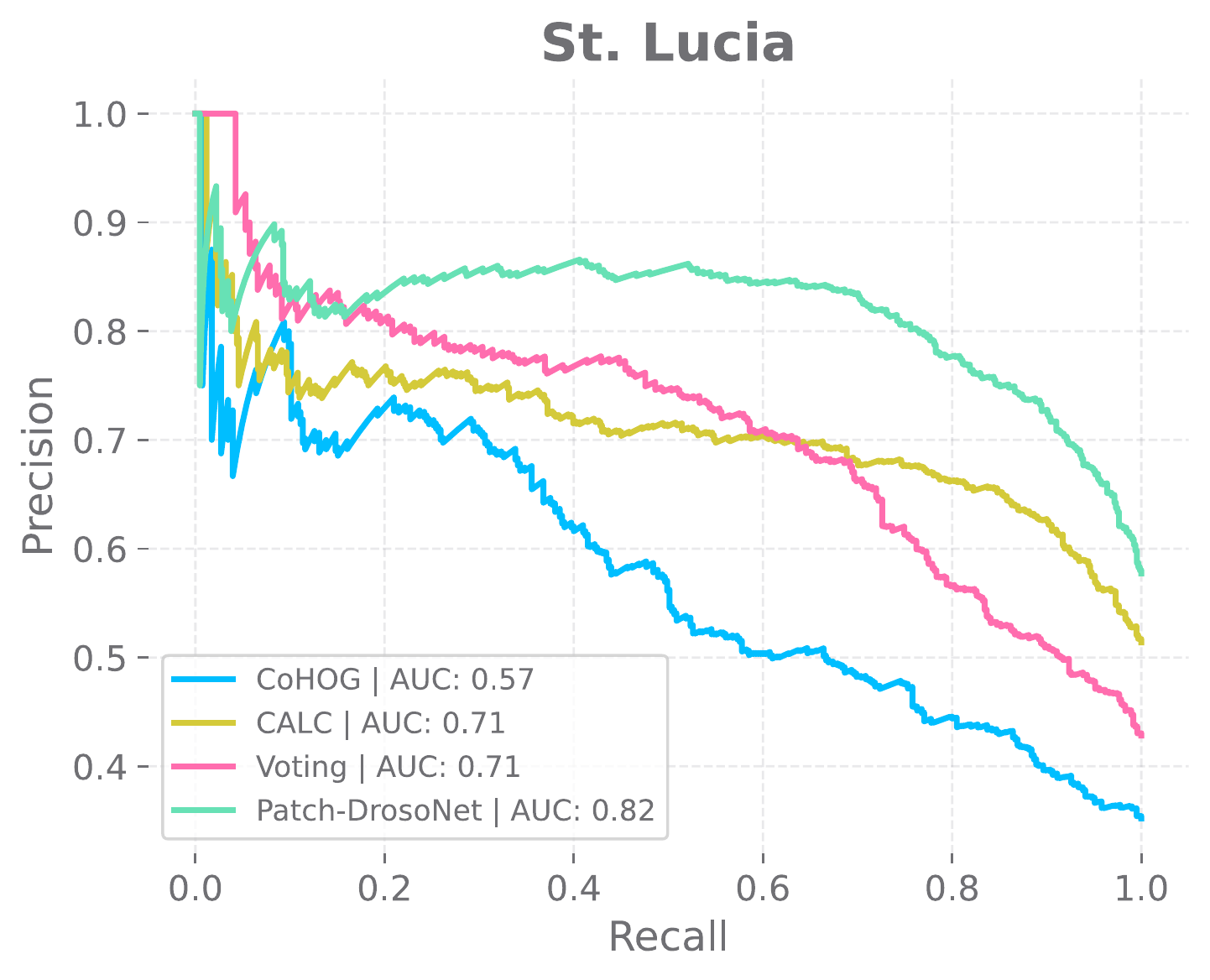}
		\caption{}
		\label{fig:pr_curves:D}
	\end{subfigure}
	\hfill
	\caption{Precision-Recall Curves}
	\label{fig:pr_curves}
\end{figure*}

Assessing VPR performance, we show how the different lightweight techniques compare in the precision-recall curves in Fig. \ref{fig:pr_curves}, alongside the respective area under these curves (AUC). In the challenging appearance changes of Nordland Winter, our proposed Patch-DrosoNet is by far the best performing method, with more than double the AUC of all other techniques. In the more moderate appearance changes presented in Fall, our method is tied with the previously established Voting system, both being the top performers. In the viewpoint lateral shift assessment of Day-Right, Patch-DrosoNet is once again tied with the legacy Voting, with CoHOG achieving the best performance. Finally, our proposed approach is again the top performer in the St. Lucia dataset, displaying improved resilience against illumination changes and dynamic elements.

\begin{figure}[thpb]
\vspace*{1ex}
\centering
\includegraphics[width=0.8\columnwidth]{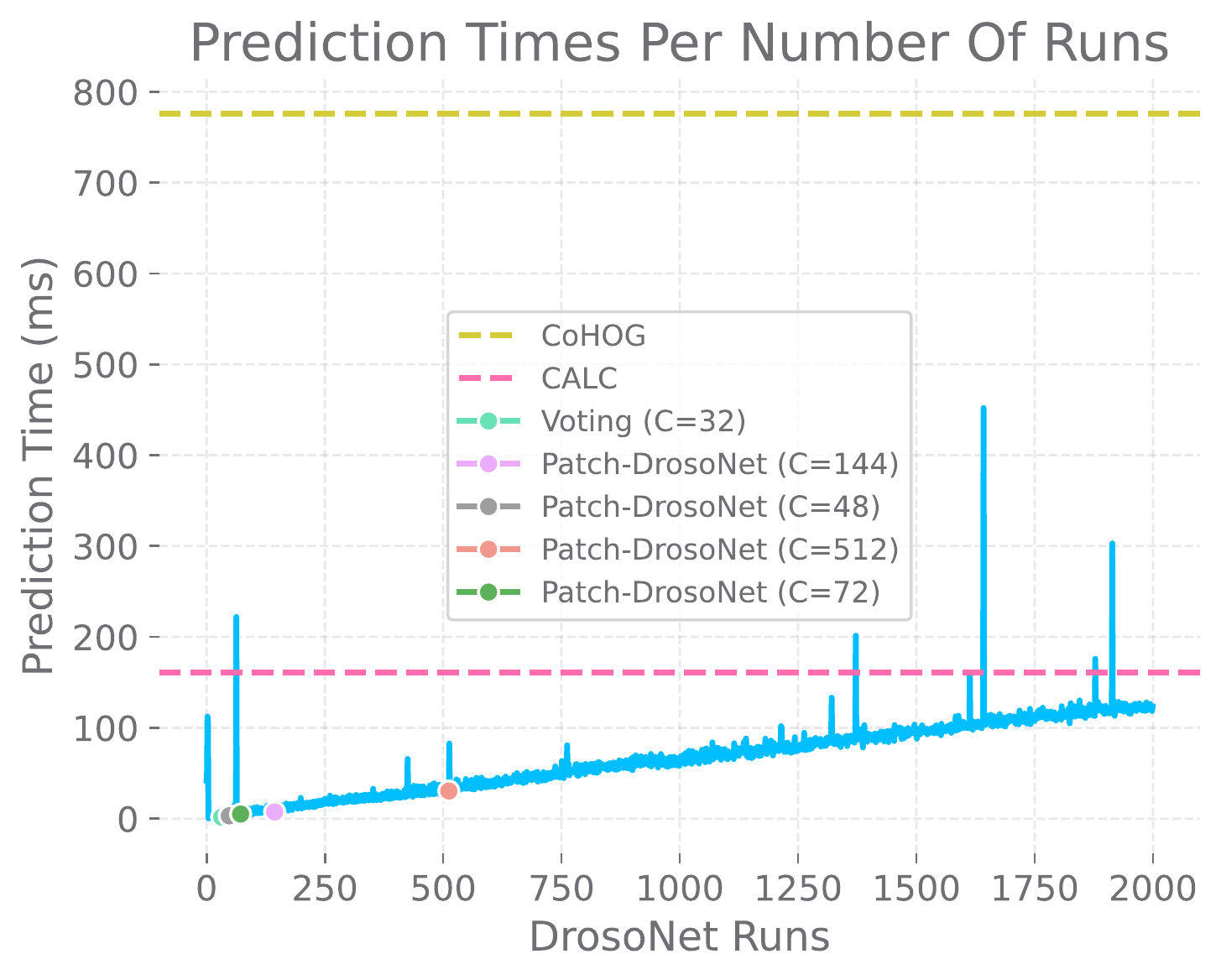}
\caption{Prediction Computation Time Comparison.}
\label{fig:comp_times}
\end{figure}

In Fig. \ref{fig:comp_times} we show the prediction times of the different techniques on a dataset with 1000 reference images. Patch-DrosoNet is the second fastest algorithm, even in the most expensive configuration tested ($C=512$ for the St. Lucia dataset). Only Voting remains faster due to a lower number of DrosoNet runs, but also present overall worse VPR performance, especially in the Winter dataset. Even with $C=2000$, Patch-DrosoNet remains faster than CALC and CoHOG, leaving open the possibility for more complex DrosoNet grouping schemes.

% \begin{table}[t]
%   \centering
%   \caption{Prediction computation times and memory requirements}
%   \label{memory&time}%
%     \begin{tabular}{M{1.5cm}M{1.5cm}M{1.5cm}}
%     \toprule
% Model     & Prediction time (ms)  & Size (MBs) \\
% \midrule
% HybridNet & 1143.92                & 61.44      \\
% CoHOG     & 3627.18               & 123.01     \\
% CALC      & 73.62                 & 4.26       \\
% HOG       & 208.71                & 142.88     \\
% FlyNet    & 1.00                   & 0.26       \\
% DrosoNet  & 1.00                  & 0.19       \\
% Voting    & 18.43                 & 6.19      \\
% Patch-DrosoNet & & \\

% \bottomrule
% \end{tabular}
% \end{table}%

\section{Conclusions and Future Work}
\label{conclusions}
In this work, we propose a novel approach to train and utilize the established DrosoNet algorithm, resulting in a lightweight VPR technique with advantages over previous work. The core of the method consists on dividing images into patches and training different DrosoNet groups to specialize on different image regions, increasing differentiation between DrosoNets. At match time, each DrosoNet outputs its scores for each region of the query image and all score vectors are then combined using the voting mechanism.

However, the system is not without limitations. The optimal image grid and number of DrosoNets varies across different datasets, making it harder to produce a more general solution. For improving this work, we suggest focusing on how to identify key regions within an image, or using a segmentation algorithm to extract different classes of patches that can then be used for specialization.

%\addtolength{\textheight}{-12cm} % This command serves to balance the column lengths

%%%%%%%%%%%%%%%%%%%%%%%%%%%%%%%%%%%%%%%%%%%%%%%%%%%%%%%%%%%%%%%%%%%%%%%%%%%%%%%%

%%%%%%%%%%%%%%%%%%%%%%%%%%%%%%%%%%%%%%%%%%%%%%%%%%%%%%%%%%%%%%%%%%%%%%%%%%%%%%%%

%%%%%%%%%%%%%%%%%%%%%%%%%%%%%%%%%%%%%%%%%%%%%%%%%%%%%%%%%%%%%%%%%%%%%%%%%%%%%%%%

\bibliographystyle{IEEEtran}
\typeout{}
\bibliography{ref}

\end{document}